\documentclass[letterpaper, 10 pt, journal, twoside]{IEEEtran} 

\usepackage{amsmath} % cmex10
\usepackage{bm}
\usepackage{amssymb}
\usepackage{siunitx}
\usepackage[pdftex]{graphicx}
\usepackage{epstopdf}
\usepackage{float}
\usepackage{cite}
\usepackage{color}
\usepackage{cancel}
\usepackage{dblfloatfix}
\usepackage{soul}
\usepackage{multirow}
\IEEEoverridecommandlockouts                              % This command is only needed if 
\newcommand{\ignore}[1]{}

\newcommand{\norm}[1]{\left\Vert#1\right\Vert} % Norm
\newcommand{\bbm}{\begin{bmatrix}}
\newcommand{\ebm}{\end{bmatrix}}
\newcommand{\bma}[1]{\left[\begin{array}{#1}}
\newcommand{\ema}{\end{array}\right]}

\DeclareMathAlphabet{\mbf}{OT1}{ptm}{b}{n}
\newcommand{\mbs}[1]{{\boldsymbol{#1}}}
\newcommand{\mbfdot}[1]{{\dot{\mbf{#1}}}}
\newcommand{\mbfbar}[1]{{\bar{\mbf{#1}}}}
\newcommand{\mbfhat}[1]{{\hat{\mbf{#1}}}}

\newcommand{\mbfcheck}[1]{{\check{\mbf{#1}}}}

\def\fdotb{{\raisebox{-0.6ex}{ \kern0.2ex\raisebox{0.8ex}{\tiny $\hspace*{-1ex}\circ$}}}}
\def\fddotb{{\raisebox{-0.6ex}{ \kern0.2ex\raisebox{0.8ex}{\tiny $\hspace*{-1ex}\circ\circ$}}}}

\newcommand{\trans}{{\ensuremath{\mathsf{T}}}} % transpose
\newcommand{\utimes}{ {\raisebox{-0.6ex}{ \kern-1.0ex\raisebox{0.6ex}{ \small $\mathsf{v}$}}} } % 
\newcommand{\beq}{\begin{equation}}
\newcommand{\eeq}{\end{equation}}
\newcommand{\bdis}{\begin{displaymath}}
\newcommand{\edis}{\end{displaymath}}
\newcommand{\beqarray}{\begin{eqnarray}}
\newcommand{\eeqarray}{\end{eqnarray}}
\newcommand{\beqarraynn}{\begin{eqnarray*}}
\newcommand{\eeqarraynn}{\end{eqnarray*}}

\renewcommand{\theenumii}{\arabic{enumii}}

\makeatletter
\renewcommand{\p@enumii}{\theenumi.}

\author{Daniil Lisus$^{1}$, Charles Champagne Cossette$^{1}$, Mohammed Shalaby$^{1}$, and James Richard Forbes$^{1}$%
\thanks{Manuscript received: February, 24, 2021; Revised June, 17, 2021; Accepted July, 18, 2021.}%Use only for final RAL version
\thanks{This paper was recommended for publication by Editor Sven Behnke upon evaluation of the Associate Editor and Reviewers' comments.
This work was supported by the Canadian Foundation for Innovation, FRQNT Team Grant, NSERC Discovery Grant, and  McGill University's William Dawson Scholar program.} %Use only for final RAL version
\thanks{$^{1}$All authors are with the Department of Mechanical Engineering, McGill University, Canada.
        {\tt\footnotesize daniil.lisus@mail.mcgill.ca}}%
\thanks{Digital Object Identifier (DOI): see top of this page.}
}
\title{Heading Estimation Using Ultra-Wideband Received Signal Strength and Gaussian Processes}

\begin{document}
 %auto-ignore
% This is not a standalone latex document. To use this file
% as a cover page on an arXiv upload of a document that is 
% already accepted as some sort of IEEE publication, you must
%
%  1) add the following just after the \begin{document} line
%     of your main paper document
%
%         \input{arxiv-cover-ieee.tex}
%
%  2) and replace the relevant information in the block below.
%
% The relevant information has been parameterized as variables.
% Simply replace the variable values with your stuff and the 
% result should be good.
%
% Make sure to not include this file for ACTUAL submissions to 
% the IEEE. Luckily you can just comment in/out the 
% \input{arxiv-cover-ieee.tex} line.
%
% FYI: The exact citation with formatting can be obtained 
% from your paper's page on IEEE Xplore.
%
%%%%%%%%%%%%%%%%%%%%%%%%%%%%%%%%%%%%%%%%%%%%%%%%%%%%%%%%%%%%%%%
%%%%%%%%%%%%%%%%%%%%%% ADD YOUR INFO HERE %%%%%%%%%%%%%%%%%%%%%
%%%%%%%%%%%%%%%%%%%%%%%%%%%%%%%%%%%%%%%%%%%%%%%%%%%%%%%%%%%%%%%
\def \myJournal {IEEE Robotics and Automation Letters}
\def \myDoi {10.1109/LRA.2021.3102300}
\def \myPaperSiteName {IEEE Xplore}
\def \myPaperSiteLink {https://ieeexplore.ieee.org/document/9508865}
\def \myYear {2021}

\def \myPaperCitation{D. Lisus, C. C. Cossette, M. Shalaby and J. R. Forbes, ``Heading Estimation Using Ultra-Wideband Received Signal Strength and Gaussian Processes,'' in \textit{IEEE Robotics and Automation Letters}, vol. 6, no. 4, pp. 8387-8393, Oct 2021.}

%%%%%%%%%%%%%%%%%%%%%%%%%%%%%%%%%%%%%%%%%%%%%%%%%%%%%%%%%%%%%%%
%%%%%%%%%%%%%%%%%%%%%%%%%%%%%%%%%%%%%%%%%%%%%%%%%%%%%%%%%%%%%%%

\begin{figure*}[t]

\thispagestyle{empty}
\begin{center}
\begin{minipage}{6in}
\centering
This paper has been accepted for publication in \emph{\myJournal}. 
\vspace{1em}

This is the author's version of an article that has, or will be, published in this journal or conference. Changes were, or will be, made to this version by the publisher prior to publication.
\vspace{2em}

\begin{tabular}{rl}
DOI: & \myDoi\\
\myPaperSiteName: & \texttt{\myPaperSiteLink}
\end{tabular}

\vspace{2em}
Please cite this paper as:

\myPaperCitation

\vspace{15cm}
\copyright \myYear \hspace{4pt}IEEE. Personal use of this material is permitted. Permission from IEEE must be obtained for all other uses, in any current or future media, including reprinting/republishing this material for advertising or promotional purposes, creating new collective works, for resale or redistribution to servers or lists, or reuse of any copyrighted component of this work in other works.

\end{minipage}
\end{center}
\end{figure*}
\newpage
\clearpage
\pagenumbering{arabic} 

\maketitle

\begin{abstract}
It is essential that a robot has the ability to
determine its position and orientation to execute
tasks autonomously. Heading estimation is especially challenging in indoor environments where magnetic distortions make magnetometer-based heading estimation difficult. Ultra-wideband (UWB) transceivers are common in indoor localization problems. This letter experimentally demonstrates how to use UWB range and received signal strength (RSS) measurements to estimate robot heading. The RSS of a UWB antenna varies with its orientation. As such, a Gaussian process (GP) is used to learn a data-driven relationship from UWB range and RSS inputs to orientation outputs. Combined with a gyroscope in an invariant extended Kalman filter, this realizes a heading estimation method that uses only UWB and gyroscope measurements. 
\end{abstract}

% Keywords appear just beneath the abstract. Use only for final RAL version.  

%%%%%%%%%%%%%%%%%%%%%%%%%%%%%%%%%%%%%%%%%%%%%%%%%%%%%%%%%%%%%%%%%%%%%%%%%%%%%%%%
\section{Introduction}
% Drop letter for first word of the Introduction
\IEEEPARstart{E}{xecution} of robotic tasks relies heavily on the ability to estimate the position and orientation of the robot. In outdoor environments, GPS is a reliable positioning solution. In indoor environments, ultra-wideband (UWB) transceivers are frequently used for positioning \cite{Sahinoglu2006}. However, the task of attitude estimation, particularly in indoor applications, is still an ongoing problem. Traditional methods use a magnetometer to predict a robot's orientation based on the local magnetic field. However, this field can be greatly distorted in indoor settings by metallic objects. Another approach, referred to position-aided inertial navigation, involves the use of inertial measurement unit (IMU) and position measurements to solve for the robot's orientation \cite[Chapter~11]{farrell2008aided}. However, heading becomes unobservable and starts to drift during motion with constant linear velocity, including when the robot is stationary or rotating in a fixed position \cite{ins_obsv}. In this letter, it is experimentally shown that UWB transceivers, commonly used to estimate position in indoor applications, can also be used to provide heading estimates.

UWB transceivers are attractive due to their low power, small size, low cost, decimeter-level ranging accuracy, and ability to act as a communication medium as well as a positioning device \cite{Ledergerber2017},  when a system of static transceivers, called anchors, is present.  The large bandwidth of UWB allows the transceivers to send nanosecond-duration radio impulses,  resulting in an increased time resolution and resistance to noise caused by reflected signals \cite{Zwirello2012}. 

\begin{figure}[!t]
	\centering
    \includegraphics[trim=0 0 0 0, clip,width=0.38\textwidth]{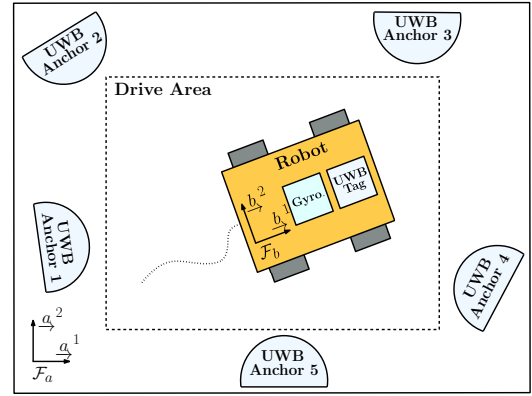}
    \caption{A sample room setup of a robot equipped with a UWB tag and gyroscope and 5 stationary UWB anchors. The local frame $\mathcal{F}_a$ and robot body frame $\mathcal{F}_b$ are shown.}
    	\label{fig:room}
    	\vspace{-0.34cm}
\end{figure}	

\begin{figure}[!b]
	\vspace{-0.3cm}
	\centering
    \includegraphics[width=0.18\textwidth]{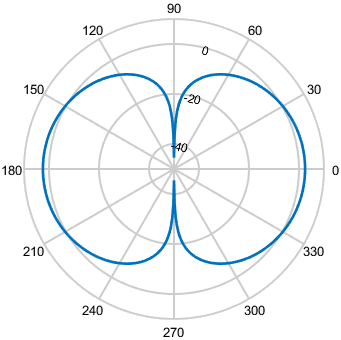}
    \includegraphics[width=0.18\textwidth]{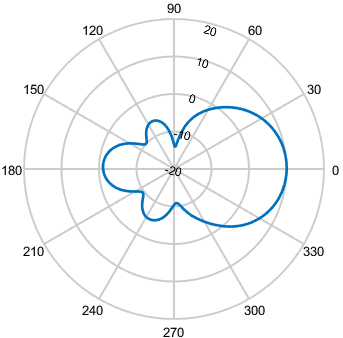}
    \caption{A simulated radiation pattern for an ideal dipole antenna (left) and an ideal helical antenna (right). The directivity value, proportional to the RSS, in dBi is plotted as a function of the angle around the antenna.}
    	\label{fig:dir_plot}
\end{figure}

Recently,  several researchers have aimed to improve the quality of UWB range measurements by utilizing Gaussian Process (GP) models. These models have been used to correct for discrepancies caused by nonuniform radiation patterns of the UWB antenna and by non-line-of-sight communications \cite{Ledergerber2017, Savic2016, Ledergerber2018, Zhao2020}. Since range measurements are used to estimate position, improved range measurements result in improved positioning accuracy. Existing literature focuses on the single-tag, multi-anchor approach where a single UWB module, called a tag, is mounted on a moving robot and measures the range to multiple stationary UWB modules, called anchors, with known positions. A depiction of this approach is shown in Figure \ref{fig:room}. Previous papers, such as \cite{Ledergerber2019, Dotlic2018}, have investigated predicting the angle of arrival of a received UWB signal. An approach for 2D pose estimation is shown in \cite{Ledergerber2019}. However, the method relies on extracting the channel impulse response, which is not possible with some off-the-shelf UWB modules. Attitude estimation based on GPS received signal strength (RSS) is shown in \cite{Gross2016}.

Compared to previous papers that use GP models to improve the accuracy of UWB range measurements, this letter explores how to use a GP model, trained using range and RSS inputs and ground truth heading outputs, to provide direct heading estimation of an indoor planar robot. The GP model's ability to ``learn" a relationship between range, RSS, and heading is due to the heading-dependent RSS pattern associated with a UWB antenna \cite{Chen2018}. This dependency is visualized in Figure \ref{fig:dir_plot} for simulated ideal dipole and helical antennas. The pattern is unique for a given antenna and independent of the transceiver. 

\begin{figure}[!t]
	\centering
    \includegraphics[trim=0 0 0 0, clip,width=0.45\textwidth]{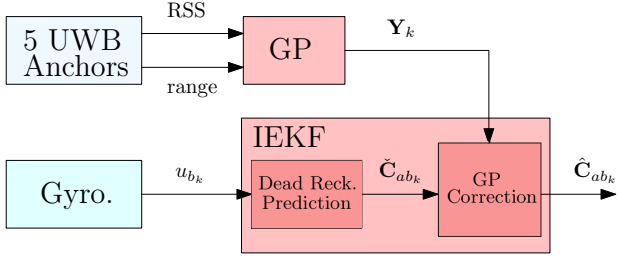}
    \caption{A high-level diagram of the proposed approach. The Gaussian Process (GP) block is pre-trained for a given anchor setup.}
    	\label{fig:approach}
    	\vspace{-0.4cm}
\end{figure}

The GP model is capable of providing both a heading estimate and a corresponding uncertainty. However, to further enhance heading estimation, the GP model is used in the correction step of an invariant extended Kalman filter (IEKF) \cite{Barrau2017} that uses a gyroscope, another common sensor, within a prediction step. The proposed approach is capable of heading estimation using only a UWB module and a gyroscope, both of which are typically present in indoor setups that localize using an IMU and range measurements. A visualization of the proposed approach is shown in Figure \ref{fig:approach}.

The rest of the letter is structured as follows. Section \ref{sec:prelim} provides a general overview of GP models and the IEKF. Section \ref{sec:prob_def} formulates the state to be estimated. Section~\ref{sec:GP_impl} describes the GP implementation used to produce a heading prediction and Section \ref{sec:filter} derives the IEKF filter that couples gyroscope data and the GP model. Finally, Section~\ref{sec:results} presents the performance of the proposed method on real data and Section \ref{sec:conclusion} provides concluding remarks.

\section{Preliminaries}\label{sec:prelim}
\subsection{Gaussian Process}
A GP models a distribution over functions, which can be viewed as an infinite collection of random variables (RVs) \cite{Rasmussen}. Specifically, GPs attempt to capture the distribution over possible functions $f(\mbf{x})$, provided with inputs $\mbf{x} \in \mathbb{R}^d$ and outputs $y \in \mathbb{R}$. In reality, the measured outputs $y$ are noisy and assumed to take the form
\begin{equation}
	y = f(\mbf{x}) + \epsilon,
\end{equation}
where the noise $\epsilon \sim \mathcal{N}(0, \sigma^2)$ is normally distributed with variance $\sigma^2$. A GP is completely defined by a mean $m\left(\mbf{x} \right)$ and covariance function or \emph{kernel} $k \left(\mbf{x}, \mbf{x}^\prime \right)$, which are explicitly
 stated as
	\begin{align}
		m \left( \mbf{x} \right) & = \mathrm{E} [ f \left( \mbf{x} \right) ],\\
		k \left( \mbf{x}, \mbf{x}^\prime \right) & =  \mathrm{E} [ \left( f ( \mbf{x} ) - m (\mbf{x} ) \right) \left( f( \mbf{x}^\prime ) - m (\mbf{x}^\prime ) \right) ].
	\end{align}
Using these expressions, a GP can be written as
	\begin{align}
		f \left( \mbf{x} \right) \sim \mathcal{GP} \left(m (\mbf{x} ), k (\mbf{x}, \mbf{x}^\prime) \right),
	\end{align}
where the distribution of the output of the function $f(\mbf{x})$ is estimated for a given input $\mbf{x}$. Typically, this value is a scalar and multiple GPs are built to model a vector of output values. The construction of multiple-output GPs greatly increases the complexity and is an active area of research \cite{Wang2015}.

A GP model produces predictions by conditioning the query RV $y^\star$ corresponding to the test input $\mbf{x}^\star \in \mathbb{R}^d$, on all of the available $n$ training RVs, which are the set of measurements $\mbf{y} = \begin{bmatrix} y_1 \; y_2 \; \cdots \; y_n \end{bmatrix}^\trans \in \mathbb{R}^n$ corresponding to inputs $\mbf{X} = \begin{bmatrix} \mbf{x}_1 \; \mbf{x}_2 \; \cdots \; \mbf{x}_n \end{bmatrix}^\trans \in \mathbb{R}^{n \times d}$. The mean and covariance functions of the conditional distribution of $y^\star|\mbf{y}, \mbf{X}, \mbf{x}^\star, \mbs{\theta}$ are given by
	\begin{align} \label{eq:gp_mean_prediction}
		m(f(\mbf{x}^\star)) & = \mbf{K}(\mbf{x}^\star, \mbf{X}) \left(\mbf{K}(\mbf{X}, \mbf{X}) \right)^{-1} \mbf{y},\\ \label{eq:gp_cov_prediction}
		\textrm{cov}(f(\mbf{x}^\star))& = \mbf{K}(\mbf{x}^\star, \mbf{x}^\star) - \mbf{K}(\mbf{x}^\star, \mbf{X}) \left(\mbf{K}(\mbf{X}, \mbf{X})\right)^{-1} \mbf{K}(\mbf{X}, \mbf{x}^\star),
	\end{align}
where $\mbf{K}(\cdot, \cdot)$ is the covariance matrix, and $\mbs{\theta}$ is the collection of hyperparameters within the covariance function. The covariance matrix has the form $\mbf{K}_{ij}(\mbf{X}, \mbf{X}') = k(\mbf{X}_i, \mbf{X}_j')$ and is sized according to the number of inputs that $\mbf{X}$ and $\mbf{X}'$ have. For example, $\mbf{K}(\mbf{x}^\star, \mbf{X}) \in \mathbb{R}^{1 \times n}$.

Training a GP is the act of optimizing
\begin{equation}
\mbs{\theta}^\star = \underset{\mbs{\theta}}{\mathrm{argmax}} \; \mathrm{log} \left(p \left( \mbf{y} | \mbf{X}, \mbs{\theta} \right) \right)
\end{equation}
in order to find the hyperparameters $\mbs{\theta}^\star$ that maximize the log maximum likelihood $\mathrm{log} \left(p \left(\mbf{y} | \mbf{X}, \mbs{\theta} \right) \right)$, which is the probability that the training inputs $\mbf{X}$ produce the measured training outputs $\mbf{y}$ through the modelled functional distribution.

\subsection{Matrix Lie Groups}\label{sec:MLG}
A matrix Lie group $G$ is a set of invertible square matrices that form a group under the operation of matrix multiplication \cite{Sol}. The matrix Lie algebra associated with a matrix Lie group $G$ is denoted as $\mathfrak{g}$ and is the tangent space at the identity element on the matrix Lie group manifold. The exponential map $\exp(\cdot): \mathfrak{g} \to G$ and the logarithm map $\log(\cdot): G \to \mathfrak{g}$ provide a way to transition between these spaces. The ``vee" operator $(\cdot)^\vee: \mathfrak{g} \to \mathbb{R}^q$ and the ``wedge" operator $(\cdot)^\wedge: \mathbb{R}^q \to \mathfrak{g}$ map the matrix Lie algebra to and from the vector space $\mathbb{R}^q$, for the $q$ degrees of freedom of the matrix Lie group elements.

Using these operators, it is possible to express an error on a matrix Lie group element $\mbf{X} \in G$ as either a left or right multiplication. A \emph{left-invariant} error, which is used throughout this paper, is of the form $\exp(\delta \mbs{\xi}^\wedge) = \mbf{X}^{-1}\mbfbar{X}$, where $\mbfbar{X} \in G$ is a nominal matrix Lie group element and $\delta \mbs{\xi} \in \mathbb{R}^k$ is associated with the error $\exp(\delta \mbs{\xi}^\wedge) \in G$. An error of this form is said to be left-invariant, since $\mbf{X}$ and $\mbfbar{X}$ can can be left multiplied by another element $\mbf{X}' \in G$ without any change in the overall error expression.

\subsection{Invariant Extended Kalman Filter}
The IEKF is a variation of the extended Kalman filter (EKF) that can be used to directly estimate elements of matrix Lie groups \cite{Barrau2017}. The IEKF uses a particular error definition, which often leads to Jacobians that are state-estimate independent. Additionally, the IEKF frequently has enhanced transient performance. In the case of attitude estimation, the IEKF enables estimation of the $SO(2)$ or $SO(3)$ matrix Lie group elements directly.

The IEKF is constructed by linearizing the process and measurement models. The linearized process model is of the form
\begin{equation}\label{eq:process_goal}
	\delta{\bm{\xi}_k} = \mbf{A}_{k-1}\delta\bm{\xi}_{k-1} + \delta\mbf{w}_{k-1},
\end{equation}
where $\exp(\delta \mbs{\xi}_k^\wedge) = \mbf{X}^{-1} \mbfbar{X}$, and $\delta\mbf{w}_{k-1} \sim \mathcal{N}(\mbf{0}, \mbf{Q}_{k-1})$ is white Gaussian process noise with covariance matrix $\mbf{Q}_{k-1}$. The linearized measurement model is of the form
\begin{equation}
	\delta\bm{\xi}^y_k = \mbf{C}_k\delta\bm{\xi}_k + \mbf{M}_k\delta\mbs{\epsilon}_k,
\end{equation}
where $\delta\bm{\xi}_k^y$ is a small change in the measurement $\mbf{y}_k$, and $\delta\mbs{\epsilon}_k \sim \mathcal{N}(\mbf{0}, \mbf{R}_{k})$ is white Gaussian measurement noise with covariance matrix $\mbf{R}_{k}$.

The prediction step of the IEKF is $\mbfcheck{X}_k = \mbf{F}(\mbfhat{X}_{k-1} , \mbf{u}_{k-1})$, where $\mbf{u}_{k-1}$ is the interoceptive measurement, $\mbfhat{X}_{k-1}$ is the corrected state estimate at $t_{k-1}$, $\mbf{F}(\mbfhat{X}_{k-1}, \mbf{u}_{k-1})$ is the nonlinear process model, and $\mbfcheck{X}_k$ is the predicted state at $t_k$. The predicted covariance is 
	\begin{equation}
	\mbfcheck{P}_{k} = \mbf{A}_{k-1}\mbfhat{P}_{k-1}\mbf{A}_{k-1}^\trans + \mbf{Q}_{k-1}.
	\end{equation}
The correction step requires the Kalman gain, that being 
	\begin{equation}
	\mbf{K}_{k} = \mbfcheck{P}_{k}\mbf{C}_{k}^\trans(\mbf{C}_{k}\mbfcheck{P}_{k}\mbf{C}_{k}^\trans + \mbf{M}_{k}\mbf{R}_{k}\mbf{M}_{k}^\trans)^{-1}.
	\end{equation}
Using a left-invariant error definition, the correction step is
\begin{equation}
	\mbfhat{X}_k = \mbfcheck{X}_k\mathrm{exp}\left[-(\mbf{K}_k(\mbfcheck{X}_k^{-1}\mbf{z}_k))^\wedge\right],
\end{equation}
where $\mbf{z}_k$ is the \emph{innovation} or the difference between the measurement and the expected measurement based on the measurement model and $\mbfcheck{X}_k$. The corrected covariance is
	\begin{equation}
	\mbfhat{P}_{k} = (\mbf{1} - \mbf{K}_k\mbf{C}_{k})\mbfcheck{P}_{k}(\mbf{1} - \mbf{K}_k\mbf{C}_{k})^\trans + \mbf{K}_k\mbf{M}_{k}\mbf{R}_{k}\mbf{M}_{k}^\trans\mbf{K}_k^\trans.
	\end{equation}
The IEKF formulation is presented in a general matrix form, indicated by the bolded letters. However, when applied to an $SO(2)$ problem, as is done in Section \ref{sec:prob_def} of this paper, many elements are scalar, and are not bolded. 

\section{Problem Definition}\label{sec:prob_def}
The goal is to estimate the discrete-time matrix Lie group state
\begin{equation}
	\mbf{X}_k = \mbf{C}_{ab_k} \in SO(2),
\end{equation}
where $\mbf{C}_{ab_k}$ is the 2D attitude of the body frame $\mathcal{F}_b$ relative to the local frame $\mathcal{F}_a$. The two frames are visualized in Figure~\ref{fig:room}. The inverse of the state is simply the transpose of the element, and is written as $\mbf{X}_k^{-1} = \mbf{C}_{ab_k}^\trans$.

The state $\mbf{X}_k = \mbf{C}_{ab_k} = \exp({{\xi}_k^\theta}^\wedge) \in SO(2)$ has one degree of freedom, the ``yaw" angle $\xi_k^\theta \in (-\pi,\pi]$. The matrix Lie algebra associated with $SO(2)$ is
\begin{equation}
	\xi^{\theta^\wedge} = \begin{bmatrix}0 & -\xi^\theta\\ \xi^\theta & 0\end{bmatrix} \in \mathfrak{so}(2).
\end{equation}
The exponential map for $SO(2)$ is given by
\begin{equation}\label{eq:exp_map}
	\mathrm{exp}(\xi^{\theta^\wedge}) = \begin{bmatrix}\cos(\xi^\theta) & -\sin(\xi^\theta)\\ \sin(\xi^\theta) & \cos(\xi^\theta)\end{bmatrix} \in SO(2).
\end{equation}

\section{Gaussian Process Implementation}\label{sec:GP_impl}
In order to produce an estimate for the full $SO(2)$ element, two separate GPs are trained, one to predict $\sin(\xi^\theta)$ and another to predict $\cos(\xi^\theta)$. Estimating the $SO(2)$ element directly avoids issues arising from ambiguities and angle wrap-around. However, one cannot guarantee that the GP predictions will be restricted to $[-1, 1]$, nor that their outputs together satisfy $\sin(\xi^\theta)^2 + \cos(\xi^\theta)^2 = 1$. As such, these ``pseudo'' sine and cosine predictions of the true $\sin(\xi^\theta)$ and $\cos(\xi^\theta)$ are denoted simply as $s$ and $c$, respectively. The range and RSS measurements from each of the 5 anchors in the room are used as inputs for both GPs. A motion capture system provides ground truth heading that is used for the training outputs and for test validation.
\vspace{-0.2cm}
\subsection{Kernel Selection and Training}
Each of the GPs use the standard squared exponential (SE) kernel for the covariance function, which can be written for two arbitrary inputs $\mbf{x}$ and $\mbf{x}'$ as \cite{kernel_cookbook} 
	\begin{align} \label{eq:squared_exponential_cov_function}
		k \left(\mbf{x}, \mbf{x}' \right) = \sigma_f^2 \mathrm{exp} \left(-\frac{1}{2 \sigma_\ell^2} \norm{\mbf{x} - \mbf{x}'}^2 \right),
	\end{align}
where $\sigma_\ell$ and $\sigma_f$ are the hyperparameters to be optimized and are known as the \emph{characteristic lengthscale} and \emph{signal variance}, respectively \cite[Chapter~2.3]{Rasmussen}. The hyperparameters are optimized using the MATLAB \texttt{fitrgp} library, which selects hyperparameters that minimize the five-fold cross-validation loss using the training data \cite[Chapter~5.3]{Rasmussen}. The SE kernel is chosen after experimentation due to its simplicity and comparable-or-better performance compared to other standard kernels, such as the Matern or ARD kernels.
\vspace{-0.2cm}
\subsection{Gaussian Process Output Normalization}
The full $SO(2)$ element can be assembled from the individual $s$ and $c$ components through the relationship \eqref{eq:exp_map}. However, since there is no guarantee that the produced element would satisfy the constraint $s^2 + c^2 = 1$, the outputs of the two GPs are first normalized and then assembled into a proper $SO(2)$ element using
\begin{equation}\label{eq:projection}
	\mbf{Y}_k = \frac{1}{\sqrt{s_k^2 + c_k^2}}\begin{bmatrix}c_k & -s_k\\ s_k & c_k\end{bmatrix}  \in SO(2),
\end{equation}
where $s_k$ and $c_k$ are GP predictions at timestep $t_k$.

The normalized prediction $\mbf{Y}_k$ is a nonlinear function of the independently predicted sine and cosine terms, each with their respective uncertainties. As such, the uncertainty of the resulting normalized element is approximated through a standard linearization procedure. The GP outputs $s_k$ and $c_k$ are random variables with mean $\bar{s}_k$ with variance $R^s$, and mean $\bar{c}_k$ with variance $R^c$, respectively. For brevity, the subscript $k$ is omitted. Equation \eqref{eq:projection} can be written as
\begin{equation}\label{eq:simple_proj}
	\mbf{Y} = \frac{1}{\sqrt{s^2 + c^2}}\bigg(c\underbrace{\begin{bmatrix}1 & 0 \\ 0 & 1\end{bmatrix}}_{\mbf{1}} + s\underbrace{\begin{bmatrix}0 & -1 \\ 1 & 0\end{bmatrix}}_{\mbs{\Omega}}\bigg).
\end{equation}
The perturbation of the first term in \eqref{eq:simple_proj} can be approximated with a first-order expansion as
\begin{equation}
\frac{1}{\sqrt{s^2 + c^2}} \approx \underbrace{\frac{1}{\sqrt{\bar{s}^2 + \bar{c}^2}}}_{\alpha_1} + \underbrace{\frac{-\bar{c}}{(\bar{s}^2 + \bar{c}^2)^{\frac{3}{2}}}}_{\alpha_2}\delta{c} + \underbrace{\frac{-\bar{s}}{(\bar{s}^2 + \bar{c}^2)^{\frac{3}{2}}}}_{\alpha_3}\delta{s}.
\end{equation}
This expansion can be included in the general perturbation of both sides of \eqref{eq:simple_proj} as
\begin{align}\label{eq:uq_deriv}
	\mbfbar{Y}(\mbf{1} - \delta{\xi^y}^\wedge) &= (\alpha_1 + \alpha_2\delta{c} + \alpha_3\delta{s}) \left((\bar{c} + \delta c)\mbf{1} + (\bar{s} + \delta s)\mbs{\Omega}\right),\nonumber\\
	-\mbfbar{Y}\delta{\xi^y}^\wedge &= \underbrace{\left((\alpha_1 + \alpha_2\bar{c})\mbf{1} + \alpha_2\bar{s}\mbs{\Omega}\right)}_{\mbf{D}}\delta c \nonumber\\ & \qquad \qquad +\underbrace{\left((\alpha_1 + \alpha_3\bar{s})\mbs{\Omega} + \alpha_3\bar{c}\mbf{1}\right)}_{\mbf{E}}\delta s,\\
	\delta{\xi^y} &= \left(-\mbfbar{Y}^{-1}\mbf{D} \right)^\vee\delta c + \left(-\mbfbar{Y}^{-1}\mbf{E} \right)^\vee\delta s\nonumber.
\end{align}
The final variance $R^{\theta}$ on the normalized element $\mbf{Y}$ is then
\begin{equation}\label{eq:GP_var}
	R^{\theta} = \left(-\mbfbar{Y}^\trans\mbf{D} \right)^\vee R^{c}{\left(-\mbfbar{Y}^\trans\mbf{D} \right)^\vee}^\trans + \left(-\mbfbar{Y}^\trans\mbf{E} \right)^\vee R^{s}{\left(-\mbfbar{Y}^\trans\mbf{E} \right)^\vee}^\trans,
\end{equation}
where $\mbfbar{Y}^\trans = \mbfbar{Y}^{-1}$. This uncertainty is used in Section \ref{sec:meas_model}.

\section{Filter Formulation}\label{sec:filter}
\subsection{Process Model}
The continuous-time attitude kinematics are given by Poisson's equation
\vspace{-0.2cm}
\begin{equation}
\label{eq:poisson}
	\dot{\mbf{C}}_{ab} = \mbf{C}_{ab}\omega_b^{{ba}^\wedge},
\end{equation}
where $\omega_b^{ba} \in \mathbb{R}$ is the angular velocity of the body frame relative to the local frame, resolved in the body frame, and time dependence is dropped for notational convenience.

The interoceptive sensor in this problem, the gyroscope, measures angular velocity subject to noise as
\begin{equation}
	u_b = \omega_b^{ba} - w_b,
\end{equation}
where $w_b \sim \mathcal{N}(0, Q_c(t)\delta(t-\tau))$ is the zero-mean additive noise, $Q_c(\cdot)$ is the power spectral density, and $\delta(\cdot)$ is the Dirac delta function. Equation \eqref{eq:poisson} can thus be rewritten as
\begin{equation}
	\dot{\mbf{C}}_{ab} = \mbf{C}_{ab}(u_b + w_b)^\wedge.
\end{equation}

In order to achieve the linearized model \eqref{eq:process_goal}, the left-invariant error definition,
\begin{equation}\label{eq:ori_error_deff}
\begin{split}
	\delta\mbf{C} &= \mbf{C}_{ab}^{-1}\mbfcheck{C}_{ab} = \mbf{C}_{ab}^\trans\mbfcheck{C}_{ab},
\end{split}
\end{equation}
is used. A left-invariant error is chosen in order to be consistent with the left-invariant measurement model. The rate of change of the error \eqref{eq:ori_error_deff} is
\begin{equation}
\label{eq:ori_error}
\begin{split}
	\delta\mbfdot{C} &= \mbf{C}_{ab}^\trans\dot{\mbfcheck{C}}_{ab} + \mbfdot{C}_{ab}^\trans\mbfcheck{C}_{ab}\\
	&= \mbf{C}_{ab}^\trans\mbfcheck{C}_{ab}(u_b + \check{w}_b)^\wedge - (u_b + w_b)^\wedge\mbf{C}_{ab}^\trans\mbfcheck{C}_{ab}\\
	&= \delta\mbf{C}u_b^\wedge - (u_b + w_b)^\wedge\delta\mbf{C}.
\end{split}
\end{equation} 
Equation \eqref{eq:ori_error} can be linearized by approximating $\delta\mbf{C} \approx \mbf{1} + \delta\xi^{\theta^\wedge}$, $w_b \approx \delta w_b$ and simplifying as
\begin{equation}
\begin{split}
\delta\dot{\xi}^{\theta^\wedge} 
	&= (\mbf{1} + \delta\xi^{\theta^\wedge})u_b^\wedge - (u_b + \delta w_b)^\wedge(\mbf{1} + \delta\xi^{\theta^\wedge})\\
	&= u_b^\wedge + \delta\xi^{\theta^\wedge}u_b^\wedge - u_b^\wedge - \delta  w_b^{\wedge} - u_b^\wedge\delta\xi^{\theta^\wedge}\\
	&\approx - \delta w_b^{\wedge} + \delta\xi^{\theta^\wedge}u_b^\wedge - u_b^\wedge\delta\xi^{\theta^\wedge} \\ 
	&= - \delta w_b^{\wedge},\\
	\delta\dot{\xi}^{\theta} &= - \delta w_b.
\end{split}
\end{equation}
The final continuous-time linearized model is
\begin{equation}
\begin{split}
	\delta\dot{\xi}^{\theta} = 
	\underbrace{0}_{A_{\textrm{c}}}\delta{\xi}^{\theta} + \underbrace{-1}_{L_{\textrm{c}}}\delta w_b,
\end{split}
\end{equation}
where the subscript $c$ implies continuous time. The discrete time $A_{k-1}$ and $Q_{k-1}$ matrices needed for the discrete time form \eqref{eq:process_goal} can be computed from $A_c$, $L_c$, and $Q_c(t)$ based on \cite[Chapter~3.5.5]{farrell2008aided}. The discrete-time linearized process model, of the form \eqref{eq:process_goal}, is
\begin{equation}
\begin{split}
	\delta{\xi}^{\theta}_k = 
	\underbrace{1}_{A_{k-1}}\delta{\xi}_{k-1}^{\theta} + \delta {w_{b_{k-1}}},
\end{split}
\end{equation}
where $ \delta{w_{b_{k-1}}} \sim \mathcal{N}(0, Q_{k-1})$.

\subsection{Measurement Model}\label{sec:meas_model}
The GP runs separately from the filter and produces individual independent components of an $SO(2)$ element. These outputs are then normalized as per \eqref{eq:projection} to form a full element of the group, which is then used as a heading measurement in the correction step of the filter. The GP also produces a variance $R^\theta_k$ on this element as per \eqref{eq:GP_var}. The discrete-time measurement is then defined as
\begin{equation}
	\mbf{Y}_k = \mbf{C}_{ab_k}\mathrm{exp}({{\epsilon}_k^\theta}^\wedge),
\end{equation}
where ${\epsilon}_k^\theta \sim \mathcal{N}(0, R^\theta_k)$ and which is left-invariant according to the definition in Section \ref{sec:MLG}. This is similar to the error definition used on quaternions in \cite{Persson2012} and on $SE(3)$ matrix Lie group elements in \cite{Lenac2018}. As such, linearization can be done directly on the measurement model by rewriting both $\mbf{Y}_k$ and $\mbf{C}_{ab_k}$ in terms of the left-invariant error as
\begin{equation}
\mbfcheck{Y}_{k}\delta{\mbf{Y}_{k}}^{-1} = \mbfcheck{C}_{ab_k}\delta\mbf{C}_{k}^{-1}\mathrm{exp}({{\epsilon}^\theta_k}^\wedge),
\end{equation}
where $\mbfcheck{Y}_{k} = \mbfcheck{C}_{ab_k}$, allowing the simplification 
\begin{equation}
\delta\mbf{Y}_{k}^{-1} = \delta\mbf{C}_{k}^{-1}\mathrm{exp}({{\epsilon}^\theta_k}^\wedge).
\end{equation}
This can be linearized by substituting $\delta\mbf{Y}_{k}^{-1} \approx \mbf{1} - \delta {\xi}_k^{y^\wedge}$, $\delta\mbf{C}_{k}^{-1} \approx \mbf{1} - \delta {\xi}_k^{\theta^\wedge}$, ${\epsilon}^\theta_k \approx \delta {\epsilon}^\theta_k$, $\mathrm{exp}({{\epsilon}^\theta_k}^\wedge) \approx \mbf{1} + \delta {{\epsilon}^\theta_k}^\wedge$, and ignoring higher order terms to yield
\begin{equation}
\begin{split}
\mbf{1} - \delta {\xi}_k^{y^\wedge} &= (\mbf{1} - \delta {\xi}_k^{\theta^\wedge})(\mbf{1} + \delta {{\epsilon}^\theta_k}^\wedge),\\
  \delta {\xi}_k^{y} &= \underbrace{{1}}_{\mbf{C}_{k}}\delta {\xi}_k^{\theta} + \underbrace{-{1}}_{\mbf{M}_{k}}{\delta{\epsilon}^\theta_k}.
\end{split}
\end{equation}

The innovation in vector space can then be defined following the left error definition as
\begin{equation}
{z}_k = \mathrm{log}(\mbf{Y}_{k}^{-1}\mbfcheck{Y}_{k})^\vee,
\label{eq:innovation}
\end{equation}
where $\mbfcheck{Y}_{k} = \mbfcheck{C}_{{ab}_k}$.

\begin{figure*}[!b]
     \centering
     \minipage[t]{0.2\textwidth}
         \centering
		\includegraphics[trim=0 0 0 50,width=\textwidth]{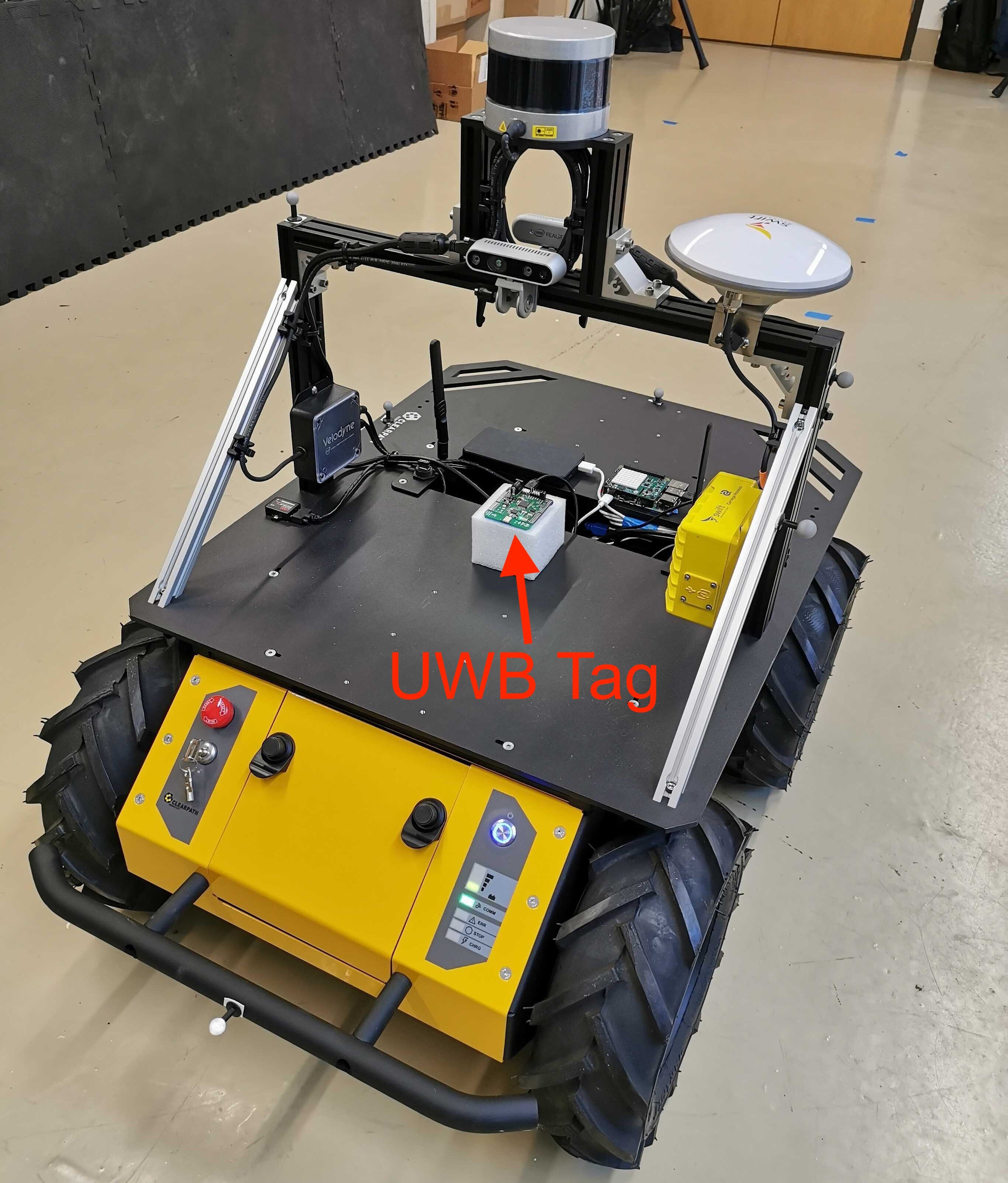}
    		\caption{The experimental setup of a Pozyx UWB Developer Tag mounted on a Clearpath Husky mobile robot.}
    		\label{fig:husky_real}
     \endminipage
     \hfill
     \minipage[t]{0.31\textwidth}
     \centering
    		\includegraphics[trim=40 10 80 70, clip,width=\textwidth]{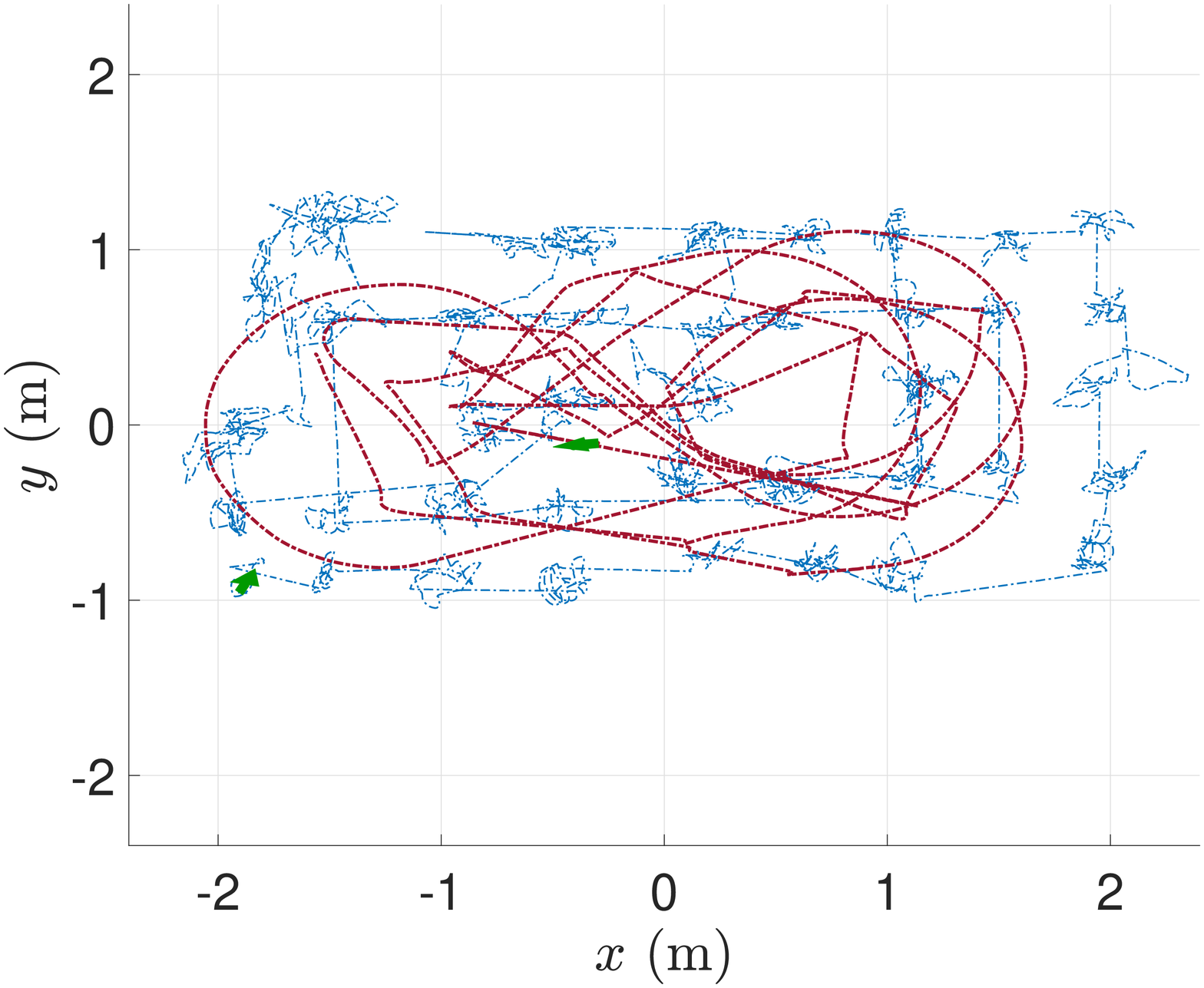}
    		\caption{The training path used for datasets Husky \#1 and \#2 (blue) and the test path of dataset Husky \#2 (red). Green arrows show the starting poses of the robot for both paths.}
    		\label{fig:trajectory}
     \endminipage
     \hfill
     \minipage[t]{0.45\textwidth}
         \centering
    \includegraphics[trim=165 10 150 30, clip,width=\textwidth]{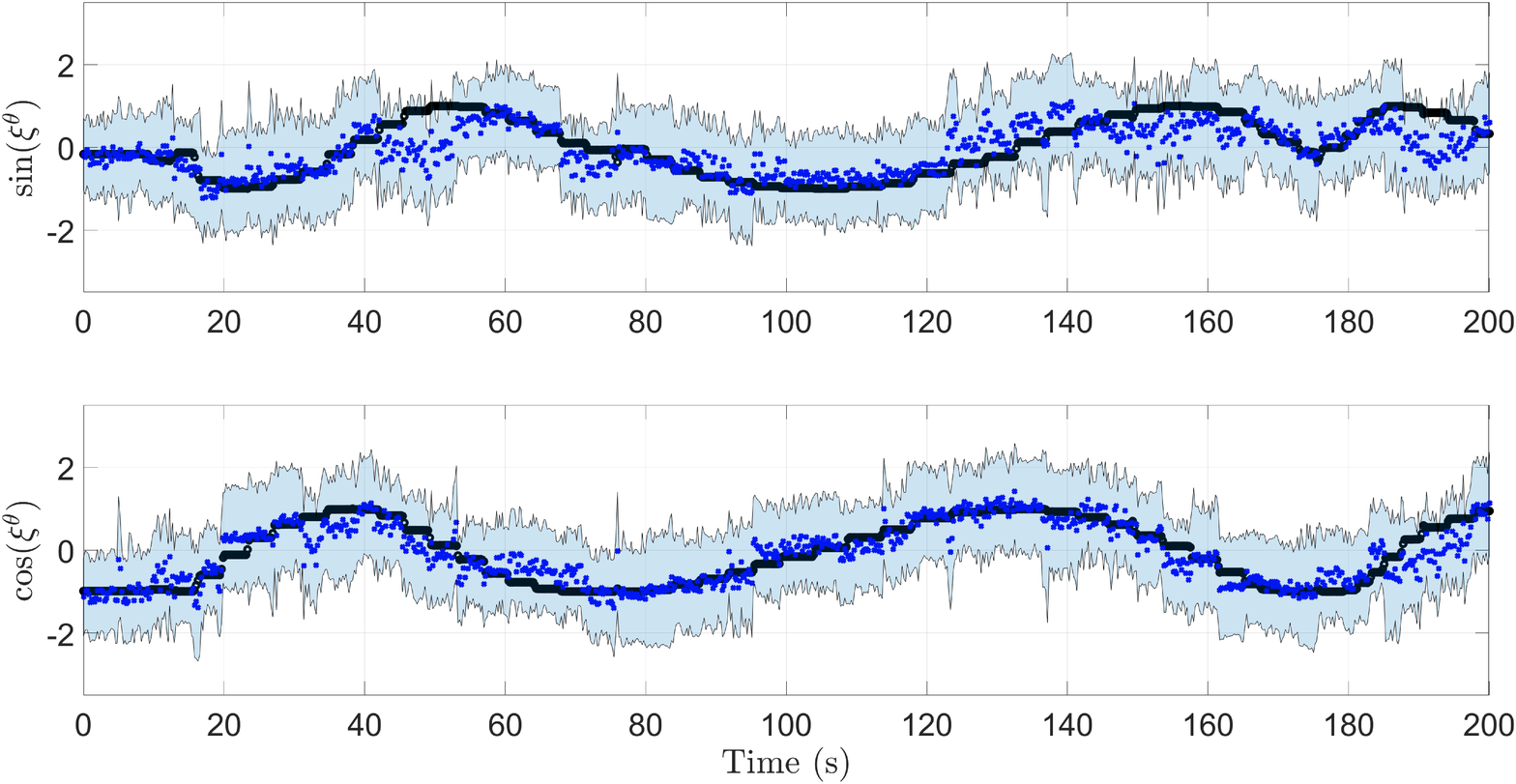}
    \caption{The ground truth $\sin(\xi^\theta)$ and $\cos(\xi^\theta)$ (black), overlayed with the GP estimates (dark blue) and their $\pm 3 \sigma$ bounds (light blue). Results are produced from the manually collected dataset, with the test set down-sampled and shortened for visual clarity.}
    	\label{fig:gp_prediction}
     \endminipage
\end{figure*}

\section{Results}\label{sec:results}
The GP and IEKF are tested on real data collected using 5 Pozyx UWB Creator Anchor modules with a Pozyx UWB Developer Tag mounted on a Clearpath Husky mobile robot, as shown in Figure \ref{fig:husky_real}. The Pozyx modules use the DW1000 transceiver and a linearly polarized UWB chip antenna. The radiation pattern of the antenna is not provided by the supplier, further motivating the use of a GP. The training data is collected by moving and rotating the robot in an area of approximately $4\; \mathrm{m} \times 2 \;\mathrm{m}$ for roughly $86,000$ data points. Two test sets are collected in the same area, with care taken to follow a path different from the training data and each other. Ground truth is recorded using a motion capture system to evaluate performance. The two test sets are referred to as the ``Husky \#1" and ``Husky \#2" datasets. The training path and one of the test paths is shown in Figure \ref{fig:trajectory}.

The results for a ``Manual" dataset are also provided for completeness. This dataset is collected by manually moving a Pozyx UWB Developer Tag in an area of approximately $2\; \mathrm{m} \times 2 \;\mathrm{m}$ for roughly $200,000$ data points. The setup for this dataset is otherwise identical to the Husky ones.

\subsection{Gaussian Process}
The numerical results of the raw GP predictions of the $SO(2)$ components $\sin(\xi^\theta)$ and $\cos(\xi^\theta)$ are shown in Table \ref{tbl:gp_results}. A visualization is shown in Figure \ref{fig:gp_prediction}. Although it is clear that a relationship is learned, the prediction quality is poor. A possible reason is that the Pozyx Developer Tags only provide RSS measurements to the nearest integer value in dBi. In these experiments, the RSS measurements vary by around $10$ dBi, meaning that only a few RSS values are fed into the GP. Thus, it is likely that the accuracy could be improved with higher-resolution RSS measurements.

\subsection{Filter}
In order to evaluate the proposed framework, three different estimators are considered.
\subsubsection{Dead-reckoning}
In order to validate the benefit of adding the GP model as a measurement step in an IEKF with a gyroscope-based prediction, a purely gyroscope-based dead reckoning estimation approach is considered.

\begin{table}[!t]
\caption{Gaussian Process prediction results.} 
\vspace{-0.3cm}
\label{tbl:gp_results}
\begin{center}
\begin{tabular}{ |c|c||c|c|c| } 
 \cline{2-5}
 \multicolumn{1}{c|}{} & Dataset & Husky \#1 & Husky \#2 & Manual\\
 \hline
 \multirow{2}{*}{$\sin(\xi^\theta)$} & RMSE & 0.52 & 0.55 & 0.49 \\ 
 & Mean $\pm 3\sigma$ & 1.35 & 1.35 & 0.79\\ 
 \hline
 \multirow{2}{*}{$\cos(\xi^\theta)$} & RMSE & 0.55 & 0.56 & 0.45\\ 
 & Mean $\pm 3\sigma$ & 1.28 & 1.28 & 0.89\\
 \hline
 \multicolumn{1}{c|}{} & No. Train Points & 17,274 & 17,274 & 25,747\\
 \cline{2-5}
 \multicolumn{1}{c|}{} & No. Test Points & 15,737 & 3,937 & 71,333\\
 \cline{2-5}
\end{tabular}
\end{center}
\vspace{-0.8cm}
\end{table} 

\subsubsection{Magnetometer Correction}
To provide a comparison to more classic indoor heading estimation methods, a magnetometer-based IEKF (mag IEKF) is also run. This filter is identical to the filter outlined in Section \ref{sec:filter}, except that the $SO(2)$ measurement in the correction step is produced from magnetometer data, as opposed to from a GP estimate. The magnetometer data is calibrated for bias and distortion from each respective training set. The local magnetic field vector is also computed based on each training set independently. The mag IEKF is tested using 100 Monte Carlo runs, where the filter is run on experimental data with a random initial prediction sampled from $\mathcal{N}({\xi^\theta_0}_{\mathrm{true}}, 1.0 \ \mathrm{rad}^2)$. The numerical results are presented in Table \ref{tbl:IEKF_results}. The poor performance on the Manual dataset is attributed to the significantly closer proximity of the UWB tag, which includes the magnetometer, to the floor and the pipes and other metallic objects below it. This likely impacted the magnetometer measurements by both causing more outliers during the test run and also decreasing the quality of the calibration procedure that was run on the training data.

\subsubsection{Gaussian Process Correction}\label{sec:IEKF_results}
An IEKF with a GP correction step, as outlined in Section \ref{sec:filter}, is tested on experimental data using 100 Monte Carlo runs with a random initial prediction sampled from $\mathcal{N}({\xi^\theta_0}_{\mathrm{true}}, 1.0 \ \mathrm{rad}^2)$. The numerical results for all datasets are presented in Table \ref{tbl:IEKF_results}. 

\begin{figure}[!t]
    \includegraphics[trim=120 10 250 50 clip,width=0.43\textwidth]{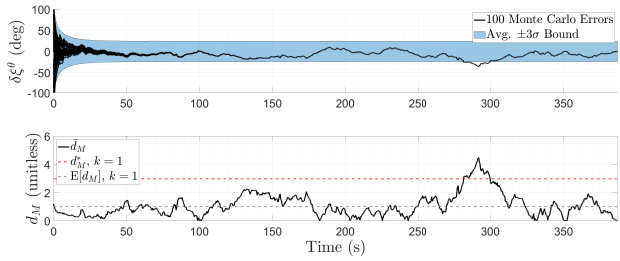}
    \caption{Top: The errors $\delta\xi^\theta$ and average $\pm 3 \sigma$ bounds of the GP IEKF estimate from 100 Monte Carlo runs. Bottom: The average Mahalanobis distance of the GP IEKF error from 100 Monte Carlo runs plotted with the $99.7\%$ 1DOF bound  (red), and the expected mean distance (grey).}
    	\label{fig:IEKF_res}
\end{figure}

For the Husky \#1 dataset, the filter produces an average RMSE of $9.74$ degrees, with an average $\pm 3 \sigma$ bound of $23.49$ degrees. The full Monte Carlo results are visualized in Figure \ref{fig:IEKF_res}. The filter is consistent and recovers from any initial prediction error. The only exception to the filter consistency occurs at approximately $292$ s, where the error and Mahalanobis distance both go outside of their respective bounds. At this time, the robot drastically changes its heading and the predictions generated using the gyroscope and the GP are both poor. Similar spikes in the error of the dead reckoning predictions can be observed in the same time period. Although the filter temporarily deviates, consistency is quickly recovered.

\begin{table}
\caption{Invariant extended Kalman filter results.}
\vspace{-0.5cm}
\label{tbl:IEKF_results}
\begin{center}
\begin{tabular}{ |c|c||c|c|c| } 
 \cline{2-5}
 \multicolumn{1}{c|}{} & Dataset & Husky \#1 & Husky \#2 & Manual\\
 \hline
 \multirow{2}{*}{GP} & $\xi^\theta$ RMSE (deg)& 9.74 & 9.16 & 11.45 \\ 
 & Steady $\pm 3\sigma$ (deg)& 23.49 & 23.49 & 22.92\\ 
 \hline
 \multirow{2}{*}{Mag} & $\xi^\theta$ RMSE (deg) & 7.90 & 9.14 & 27.61 \\ 
 & Steady $\pm 3\sigma$ (deg)& 20.93 & 20.91 & 38.37\\
 \hline
 \multicolumn{1}{c|}{} & Test Time (sec) & 386.87 & 96.42 & 786.68\\
 \cline{2-5}
\end{tabular}
\end{center}
\vspace{-0.6cm}
\end{table}

\begin{figure}[H]
	\centering
    \includegraphics[trim=120 10 180 60, clip,width=0.43\textwidth]{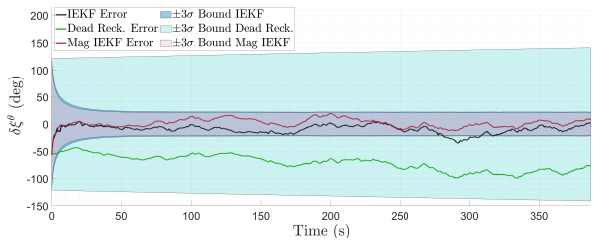}
    \caption{The error $\delta\xi^\theta$ and the $\pm 3 \sigma$ bounds produced by dead reckoning (green), the mag IEKF filter (red), and the GP IEKF filter (black) initialized with $1$ radian of error.}
    	\label{fig:dead_rec_results}
\end{figure}

\begin{figure}[!t]
	\centering
    \includegraphics[trim=130 8 170 40, clip,width=0.43\textwidth]{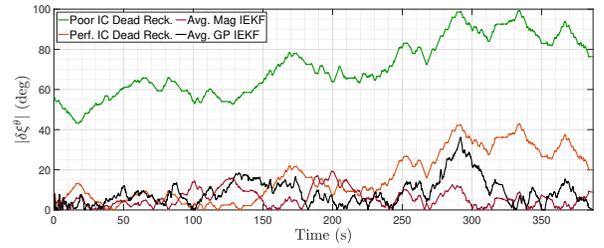}
    \caption{The absolute value of the error $\delta\xi^{\theta}$ for perfectly (orange) and imperfectly (green) initialized dead reckoning, and for 100 Monte Carlo averaged mag IEKF (red) and GP IEKF (black).}
    	\label{fig:abs_del_xi}
    	\vspace{-0.6cm}
\end{figure}

A comparison between the estimates produced on the Husky \#1 dataset by the various estimators is presented in Figure \ref{fig:dead_rec_results},  with Figure \ref{fig:abs_del_xi} showing the absolute value of the error over time. As expected, the dead reckoning prediction suffers from an inability to correct initial errors and slowly drifts even in perfectly initialized scenarios. The mag IEKF slightly outperforms the GP IEKF on the Husky datasets, although the performance is comparable. On the Manual dataset, magnetic field disturbances caused by proximity to the floor likely result in the poor performance of the mag IEKF. 

\section{Conclusion}\label{sec:conclusion}
This letter shows the feasibility of estimating the heading of a vehicle using UWB range and RSS measurements to 5 stationary anchors in a GP model. When combined with a gyroscope in an IEKF framework, the GP estimates are shown to realize convergence from arbitrary initial conditions. The filter produces a heading estimate with an RMSE of approximately $10$ degrees and a $\pm 3 \sigma$ bound of approximately $23$ degrees across three trials. It is hypothesized that with access to higher resolution RSS measurements, the GP estimate and overall heading prediction could be improved.

\bibliographystyle{IEEEtran}
\bibliography{IEEEabrv,arXiv_submission}

\end{document}